\documentclass[]{spie}  %>>> use for US letter paper
\usepackage{amsmath,amsfonts,amssymb}
\usepackage{graphicx}
\usepackage{styles/multirow}
\usepackage[colorlinks=true, allcolors=blue]{hyperref}

\title{MONCE Tracking Metrics: a comprehensive quantitative performance evaluation methodology for object tracking}

\author[a]{Kenneth Rapko}
\author[a]{Wanlin Xie}
\author[a]{Andrew Walsh}

\affil[a]{Lockheed Martin AI Center}

\authorinfo{Further author information: (Send correspondence to A.W.)\\
A.W.: E-mail: andrew.g.walsh@lmco.com}

\begin{document}
\maketitle

\begin{abstract}
    Evaluating tracking model performance is a complicated task, particularly for non-contiguous, multi-object trackers that are crucial in defense applications.  While there are various excellent tracking benchmarks available, this work expands them to quantify the performance of long-term, non-contiguous, multi-object and detection model assisted trackers.  We propose a suite of MONCE (Multi-Object Non-Contiguous Entities) image tracking metrics that provide both objective tracking model performance benchmarks as well as diagnostic insight for driving tracking model development in the form of Expected Average Overlap, Short/Long Term Re-Identification, Tracking Recall, Tracking Precision, Longevity, Localization and Absence Prediction.
\end{abstract}


\section{INTRODUCTION}
\label{sec:intro}  % \label{} allows reference to this section

Object tracking is the identification of a unique object over a contiguous time period with multi-object tracking (MOT) being the logical extension of this idea. There is a long history of different methods for evaluating the performance of single object trackers \cite{Kristan2015,Kristan2016, Wu2015} and multi-object trackers \cite{Benardin2008, Luiten2020, Luo2017}.

\subsection{Motivation}
Tracking multiple objects that are not contiguously in-frame over long time periods (and its subsequent benchmarking \cite{Valmadre2018}) is a lesser studied problem that is of particular importance to the defense industry. Current tracking performance benchmarks are largely dataset dependent or do not yet offer methods of root cause analysis for the metric score achieved. The MONCE metric suite specializes in multi-object non-contiguous entity tracking, offers root cause analysis for metric scores and is robust to sequence length variations in datasets.

\subsection{Definitions}

Before continuing onto the MONCE metrics, we would like to clarify the definitions of common tracking terms that will be used frequently in the subsequent paragraphs. An \textit{entity-frame} is an instance of an entity in a single video frame. A \textit{Unique ID (UID)} is the identifier that is given to a unique entity and should be consistent for that entity across all frames. Matching between ground truth and predicted entity-frames can be done with one of two criteria: (1) original UID or (2) any UID. \textit{Original UID} indicates that only the first predicted UID that is associated with a given ground truth UID can be matched in the frames to follow. \textit{Any UID} means that any predicted UID can be matched with a ground truth UID as long as the predicted UID has not been previously associated with a different ground truth UID. Additionally, a \textit{sequence} is defined as all frames from the initial appearance of a unique entity until the end of the video. Temporary occlusions or exits from frame will count towards the total sequence length. The necessity for this definition is that a false prediction after the ground truth entity has left is still a failure that must be captured. Any preferred method may be used to match ground truth entity-frames to predicted entity-frames. For the sake of the examples in this paper, we are using a custom matching technique that first maximizes the quantity of matches and then matches on the basis of a distance metric (intersection over union).

We also clarify definitions of specific success and failure types. Table 1 defines these conditions. Any UID and original UID matching criteria are notated with (any UID) and (original UID) respectively.

\begin{table} [ht]
    \begin{center}
     \caption{Definitions of tracking success and tracking failure.}
     \label{fig:definitions}
     \begin{tabular}{c}
     \includegraphics[scale=0.23]{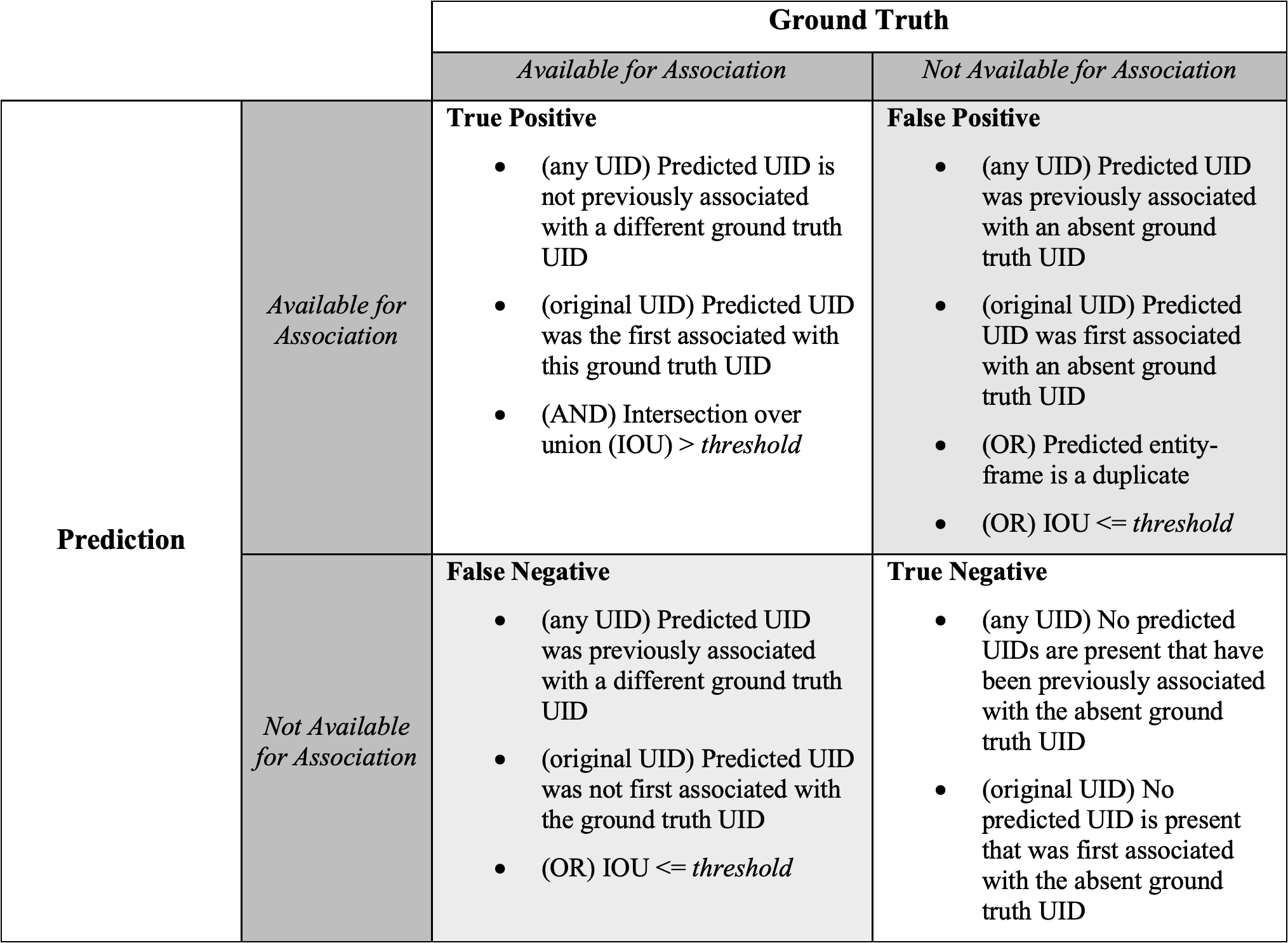}
     \end{tabular}
     \end{center}
 \end{table}

\section{MONCE METRICS SUITE METHODS}
\begin{figure} [ht]
   \begin{center}
    \begin{tabular}{c}
    \includegraphics[scale=0.5]{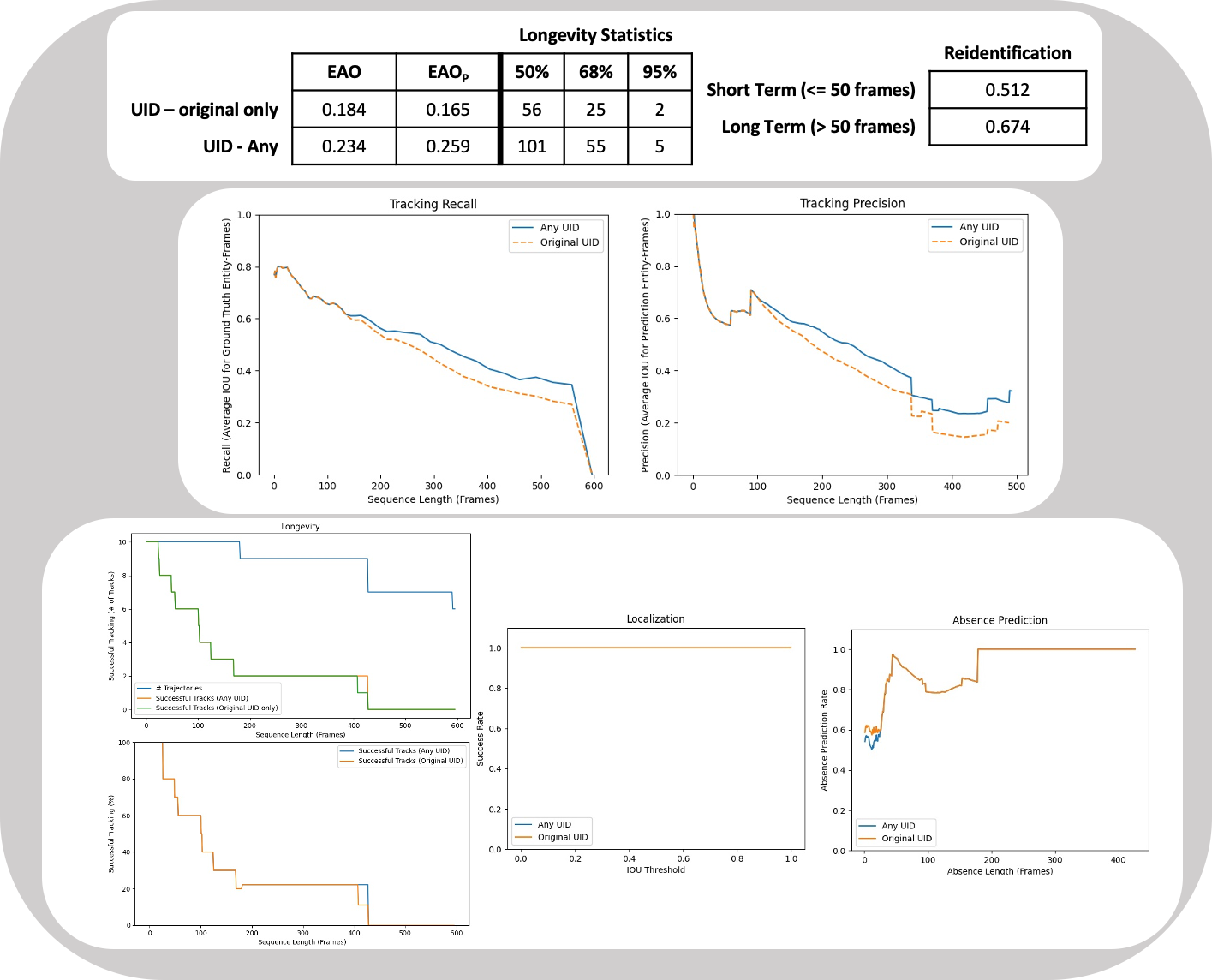}
    \end{tabular}
    \end{center}
    \caption{MONCE Dashboard for a multi-object tracker (MOT). Top row displays top-level summary metrics. Middle row displays mid-level tracking recall and precision plots. Bottom row shows specific low-level analyses to expose tracking failure modes.}
    \label{fig:dashboard}
\end{figure}

The MONCE suite includes a variety of metrics, which can be broken into 3 levels of increasingly specialized analysis. The highest, summary-level metrics are single-number scores that allow the user to assess performance immediately and evaluate performance trends between models and over time during model training and development. The mid-level metrics are plots that allow the user to examine tracking performance as a function of sequence length. The low-level encapsulates the “fine grain” metrics that allow the user to isolate specific tracking failure modes. Figure \ref{fig:dashboard} displays a dashboard of MONCE performance metrics for an arbitrary tracking model.

\subsection{Top-Level Metrics}
\label{sec:top-metrics}
MONCE offers users four high level, single number tracking performance scores: Expected Average Overlap (EAO), Expected Average Overlap Precision (EAO\textsubscript{P}), Longevity Statistics, and Short/Long Term Reidentification (REID). Single number metrics are necessary for comparative analysis across different tracking models and to view tracking performance over time for a single model. EAO is a measure of expected overlap for ground truth entity-frames and is directly taken from the VOT challenge \cite{Kristan2015}. EAO is very similar to an average of Tracking Recall \cite{Lukezic2018} with the exception that EAO assesses ground truth overlap using a subset of the sequence lengths rather than the entire dataset.

We extended the EAO concept by calculating EAO\textsubscript{P}, an average of Tracking Precision \cite{Lukezic2018} over the same subset of sequence lengths. EAO and EAO\textsubscript{P} provide single number summary metrics to analyze localization performance along with false negatives and false positives respectively. We also provide Longevity Statistics which provide the sequence length at which a given percentage of unique entities are tracked successfully: no false positives or false negatives. The final high-level metric is Short/Long Term REID, which computes the rate at which entities are successfully re-identified after leaving frame and returning.

The computation for EAO and EAO\textsubscript{P} begins with the creation of Tracking Recall and Tracking Precision curves respectively. A Kernel Density Estimate (KDE) is calculated to assess the most common range of sequence lengths \cite{Kristan2015}. The easiest solution is to use the full range of sequence lengths. However, the metric scores associated with very short and very long sequences may skew the overall result. Therefore, the KDE is used to remove these outliers. Similarly, the Longevity Statistics can be calculated by selecting a percentage success rate on the Longevity plot and reading off the appropriate sequence length. The calculation and meaning of the Longevity plot are further discussed in Section \ref{sec:low-metrics}.

A common problem in long-term tracking is the reacquisition of ground truth entities once they return to frame. To provide insight into this problem, we propose the REID rate, a metric that judges the tracking model's ability to correctly predict a true positive following an absence of a ground truth entity. To calculate the REID rate, first partition each contiguous series of absent frames into short-term or long-term absences based on a user selected threshold. If a true positive occurs immediately following an absence, assign a score of 1 for that absence. Otherwise assign a score of 0. Finally, average the scores for all short-term and long-term absences to produce their respective REID rates.

\subsection{Mid-Level Metrics}
\label{sec:mid-metrics}
The mid-level metrics consist of the Tracking Recall Plot and the Tracking Precision Plot, which allow the user to assess performance at variable sequence lengths. The EAO and EAO\textsubscript{P} metrics are heavily dataset dependent. Given two datasets with dramatically different sequence lengths, the dataset with the longer sequences will generally exhibit lower performance. After a tracking model loses track of an entity, the errors will often accumulate for the remainder of the stream and continue to drive down the EAO and EAO\textsubscript{P} scores. Longer sequences have more frames to accumulate errors, but this can be mitigated by trimming long sequences and analyzing Tracking Recall and Tracking Precision at a variety of sequence lengths. Thus, the key difference between data sets is no longer sequence length but the data quality such as brightness, resolution, contrast, or training set similarity.

The Tracking Recall plot is taken from the overlap curve in the VOT Challenge \cite{Kristan2015} with the name and calculation derived from Lukei, et al. \cite{Lukezic2018}. The Tracking Recall Plot rewards good localization and penalizes false negatives. It is an analysis of the average overlap of ground truth entity-frames as a function of sequence length. The Tracking Precision plot is computed and presented in a similar manner, using the VOT overlap curve as inspiration and the name and calculation from Lukei, et al. \cite{Lukezic2018}. The Tracking Precision Plot rewards good localization and penalizes false positives. It captures the average overlap of predicted entity-frames as a function of sequence length.

\subsection{Low-Level Metrics}
\label{sec:low-metrics}
We hope to provide users with insight into the strengths and weaknesses of their tracker and allow them to intelligently adjust their models and training approaches. In addition, these metrics isolate specific success and failure modes that contribute to the mid-level and high-level metric scores. We propose three low-level metrics: Tracking Longevity, Tracking Localization and Absence Prediction.

Tracking Longevity evaluates a tracker's ability to consistently report true positives and true negatives, independent of localization performance. To calculate Tracking Longevity at a given sequence length T, consider all ground truth sequences with a length greater than or equal to sequence length T. For each of the ground truth entity-frames within these sequences, evaluate if they have been "successfully tracked" at and up to length T \cite{Lukezic2016}. In the context of Longevity, a track is a sequence for a single ground truth entity, which may be trimmed to length T. The criterion for success for a single track is to have no false positives or false negatives with a threshold IOU of 0. For each sequence length T, we report the number of successful tracks as well as the total number of tracks in the dataset. We also display this plot as a percentage success rate for each sequence length.

Tracking Localization assesses the tracker's ability to align the predicted entity-frames with ground truth entity-frames. To calculate Tracking Localization, collect all entity-frames within successful tracks from the Longevity analysis, regardless of sequence length. Next, calculate the percentage of entity-frames that meet or exceed a series of IOU thresholds between 0 and 1. Because only successful sequences are considered, the Localization success rate is expected to be 1.0 at an IOU threshold of 0. Since perfect localization is nearly impossible, the Localization success rate is expected to approach 0 at an IOU threshold of 1.
Another common problem in long-term tracking is detecting when a ground truth entity is absent. Therefore, we analyze true negatives exclusively in the form of a metric called Absence Prediction Rate. At a given absence length T\textsubscript{a}, consider all ground truth sequences with an absence greater than or equal to T\textsubscript{a}. Trim the sequence to start at the beginning of the absence and end after T\textsubscript{a} frames. For each of these absent frames, evaluate if the tracker has correctly predicted a true negative. If a true negative was correctly predicted (i.e. no predictions), assign a score of 1. Otherwise assign a score of 0. Finally, average the scores across all of the ground truth absent frames to create an Absence Prediction rate at length T\textsubscript{a}. The resulting plot displays absence prediction rate as a function of absence length.

\section{RESULTS}
In this section, we illustrate different examples where the MONCE suite of metrics provides actionable insight. All figures were generated from real data and tracking models.

\subsection{Analyzing a tracking model in development}
The MONCE dashboard presented in Figure \ref{fig:dashboard} shows performance data for a tracker in development. The tracking model was given \textit{perfect detections}, meaning that the ground truth entity-frame locations and sizes were passed in as predicted entity-frames. Therefore, the only responsibility of the tracking model is to assign UIDs. In the case of an imperfect detection model, you can expect additional prediction errors and lower performance. We demonstrate how to utilize the MONCE metrics to characterize the weaknesses inherent in this tracker.

\begin{figure} [ht]
    \begin{center}
     \begin{tabular}{c}
     \includegraphics[scale=.9]{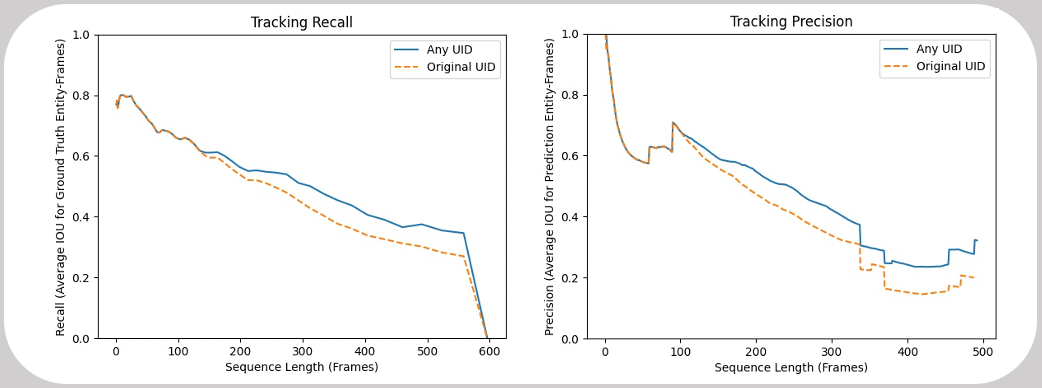}
     \end{tabular}
     \end{center}
     \caption{Tracking Recall and Tracking Precision plots for sample tracking model given perfect detections.}
     \label{fig:recall_precision}
 \end{figure}

The Tracking Recall curve in Figure \ref{fig:recall_precision} displays a "tail" on the left side of the plot. This potentially indicates that the model must observe an entity for a few frames before assigning an initial track UID. If this is unexpected, then this behavior would be a key point of investigation. We see matched performance between any UID and original UID up until around 120 frames, indicating excellent UID persistence at this sequence length. At longer sequence lengths the tracker is unable to consistently maintain the original UID and occasionally associates more than one UID with a single ground truth entity.

\begin{figure} [ht]
    \begin{center}
     \begin{tabular}{c}
     \includegraphics[scale=.9]{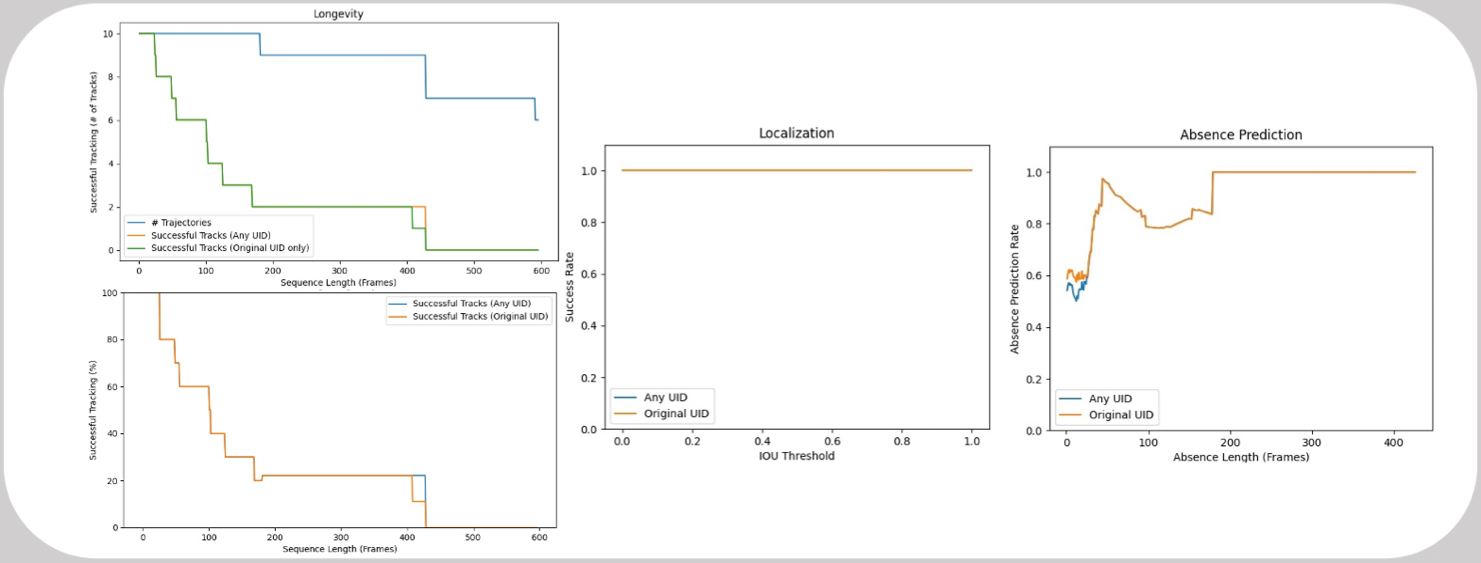}
     \end{tabular}
     \end{center}
     \caption{MONCE Tracking Longevity, Localization, and Absence Prediction plot on a sample tracker.}
     \label{fig:longevity_localization_absence}
 \end{figure}

The Tracking Longevity plot in Figure \ref{fig:longevity_localization_absence} illustrates that the tracking model demonstrates near perfect ability to classify true negatives and true positives in the first ~30 frames. This yields important information for model troubleshooting: Is the tracking model structurally inadequate for long sequences? Is it overfit to the training data? Did the training dataset not include enough sequences longer than 30 frames? Did the test dataset present challenges that were not reflected in the training dataset? Between frames 30 and 175, the increasing sequence length reveals incremental deficiencies in UID persistence.

The Localization curve is artificially constrained to a 1.0 success rate due to the perfect detection inputs.

The Absence Prediction plot in Figure \ref{fig:longevity_localization_absence} shows that it takes the tracker about 175 frames to reliably recognize that an entity has left the frame. Since a perfect detection model is used, the tracker falsely continues to assign the UID of absent ground truth entities. To improve performance on this metric, the user might consider increasing false positive penalties during training. During inference, the user might alternatively set a higher confidence threshold for assigning a UID to a track but this might exacerbate the problems with initial UID assignment.

\subsection{Recognizing a defective detection model}
\begin{figure} [ht!]
    \begin{center}
     \begin{tabular}{c}
     \includegraphics[scale=1]{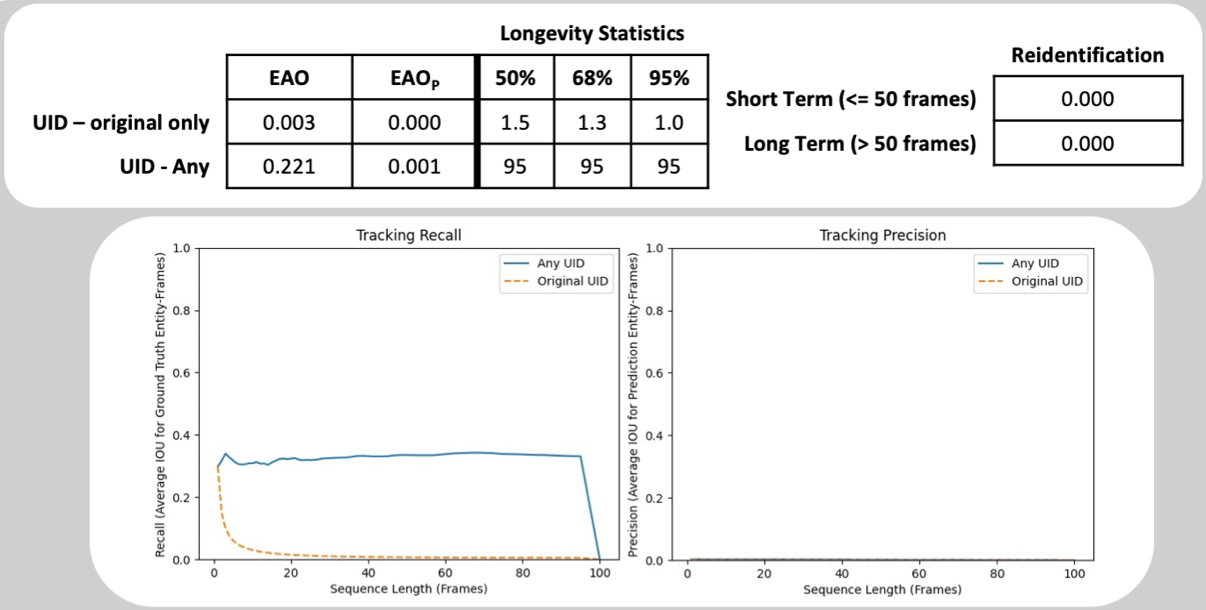}
     \end{tabular}
     \end{center}
     \caption{Sample dashboard for a detector that predicts 1000 entity-frames per frame. The predicted boxes receive a UID from 1 to N with N being the total number of predicted boxes across the entire video sequence.}
     \label{fig:defective_detection}
 \end{figure}

Another instance in which the MONCE metrics may help diagnose failures is the identification of an overactive detection model. The MONCE metrics in Figure \ref{fig:defective_detection} show performance for a detection model assisted tracker.

The user can immediately deduce that there are many false positives from the low EAO\textsubscript{P} and Tracking Precision. Additionally, the rapidly diminishing original UID curve on the Tracking Recall plot indicates that the UID assignments are unstable.

\begin{figure} [ht!]
    \begin{center}
     \begin{tabular}{c}
     \includegraphics[scale=0.3]{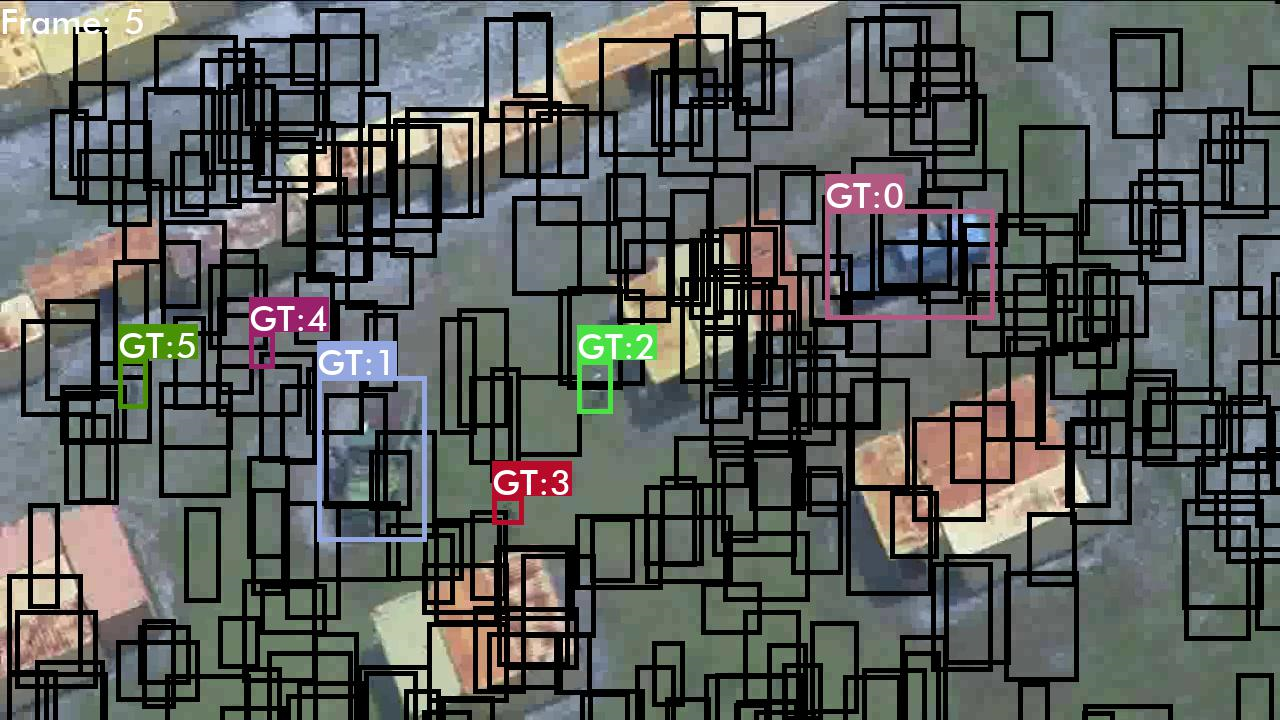}
     \end{tabular}
     \end{center}
     \caption{Sample frame with overactive detector. Ground truth entity-frames are shown in color with annotations. Predicted entity-frames are shown in black without their UIDs.}
     \label{fig:overactive_detector}
 \end{figure}

This example illustrates the advantages of analyzing Tracking Precision/EAO\textsubscript{P} alongside the more conventional Tracking Recall/EAO. Figure \ref{fig:overactive_detector} shows a single annotated output frame that supports the MONCE-derived conclusions.

\subsection{Accounting for small gaps in tracking performance}
\begin{figure} [ht!]
    \begin{center}
     \begin{tabular}{c}
     \includegraphics[scale=.9]{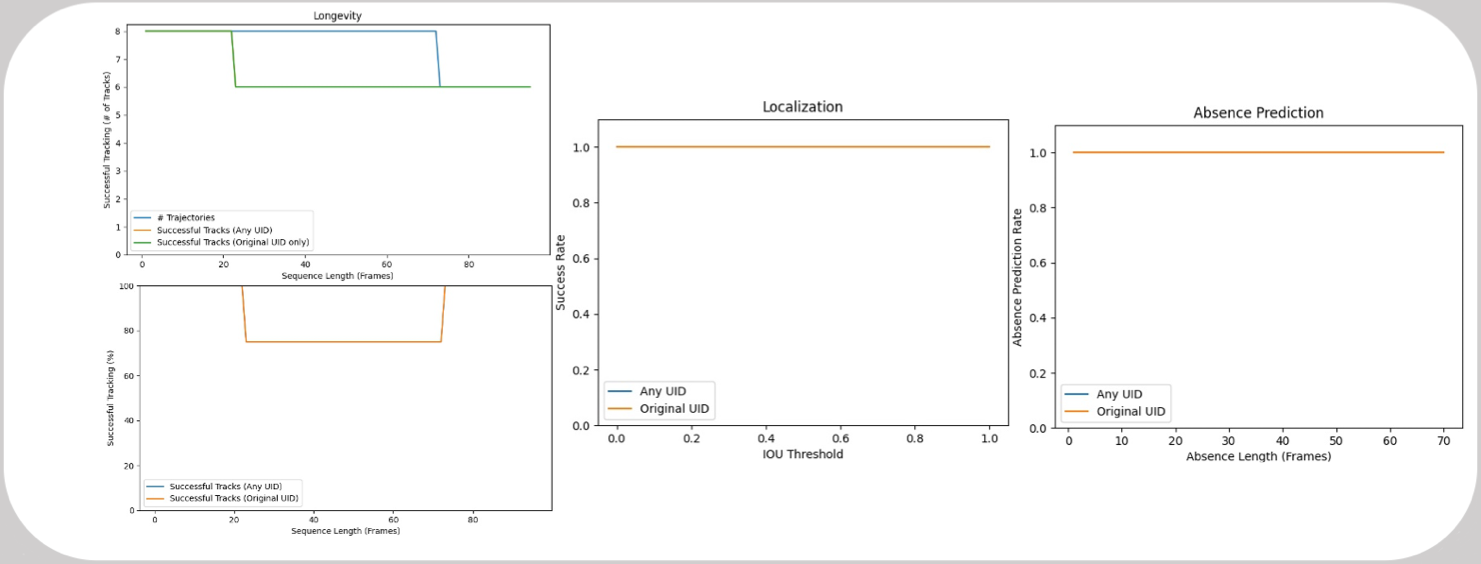}
     \end{tabular}
     \end{center}
     \caption{MONCE plots of Longevity, Localization, and Absence Prediction on an example tracking video of 100 frames. The curves for Any UID and Original UID on each graph are on top of each other (Longevity graph: orange and green lines; other graphs: blue and orange lines).}
     \label{fig:MONCE_performance_gaps}
 \end{figure}

We illustrate an example of how to utilize the MONCE metrics to pinpoint specific tracking model failures when fine-tuning tracking model performance. Figure \ref{fig:MONCE_performance_gaps} shows the low-level MONCE metrics for an object tracker that scored a .990 EAO and EAO\textsubscript{P} with perfect detections on a short 100-frame video.

\begin{figure} [ht!]
    \centering
    \begin{tabular}{ccc}
        \includegraphics[scale=1]{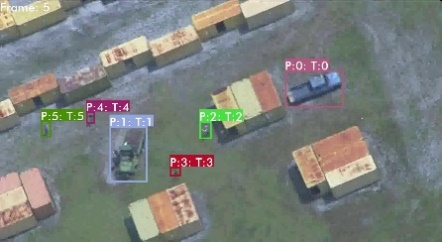} &
        \includegraphics[scale=1]{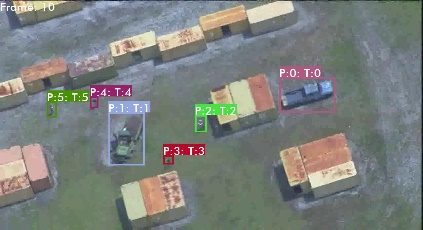} &
        \includegraphics[scale=1]{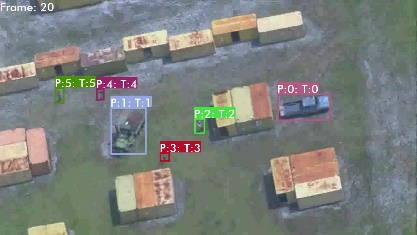} \\
        \includegraphics[scale=1]{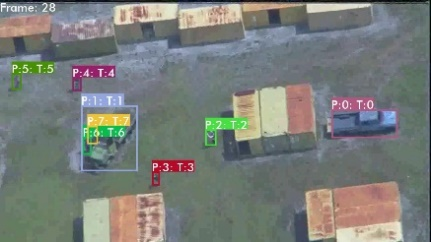} &
        \includegraphics[scale=1]{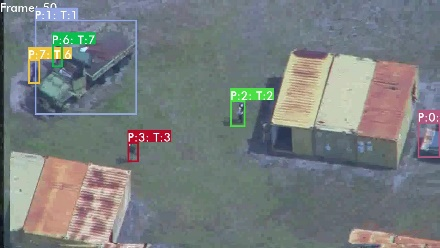} &
        \includegraphics[scale=1]{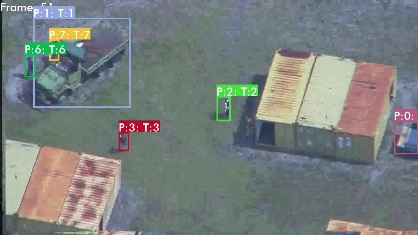} \\
    \end{tabular}

    \caption{Video stills of 8 entities moving around obstacles. The label for each entity displays its P: [predicted UID], T: [ground truth UID]. (a) Frame 5: shows 6 entities. (b,c) Frames 10 and 20: 20 frames after the appearance of these 6 entities, there are still no tracking errors, excluding them from being the cause of the longevity error (d) Frame 28: two new entities appear (6 and 7: yellow and dark green) (e) Frame 50: Around 20 frames after the appearance of UIDs 6 and 7, their predicted UIDs switch. (f) Frame 51: UIDs 6 and 7 switch back to their original assignment immediately after the error frame.}
    \label{fig:sample_video}
 \end{figure}

The Absence Prediction and the Localization graph do not show any errors. The Longevity plot indicates that all 8 entities are successfully tracked up to 20 frames, after which only 6 are successfully tracked. We can assume that those 2 entities switched their track UIDs around 20 frames. The Longevity ($\%$) plot illustrates a return to perfect tracking at about 70 frames because the two entities that exhibited errors had shorter sequence lengths. Figure \ref{fig:sample_video} analyzes the tracking output to validate the conclusions drawn from the MONCE metrics.

\subsection{Discovering incorrect labeled data}
\begin{figure} [ht!]
    \begin{center}
     \begin{tabular}{c}
     \includegraphics[scale=.5]{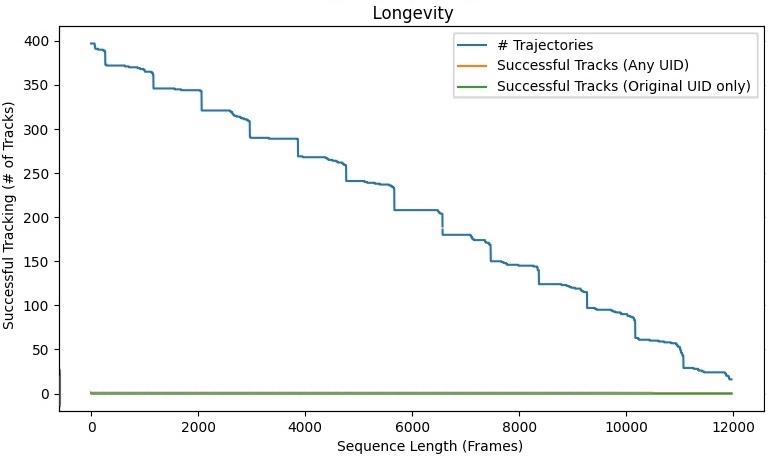}
     \end{tabular}
     \end{center}
     \caption{Longevity plot that indicates a problem with our hand labeling effort.}
     \label{fig:label_err}
 \end{figure}

The MONCE metrics have also proven to be an informative tool for adjacent development areas. For training and testing purposes, we hand labeled a tracking dataset with bounding boxes and corresponding entity-frame UIDs. After applying a detection/tracking model to the new dataset, we used the MONCE metrics to assess performance. The resulting Longevity plot is shown in Figure \ref{fig:label_err}.

At first glance it appears that our detection/tracking model scored very poorly. However, upon closer examination, the total number of trajectories (in blue) exhibits a periodic downward stairstep pattern. This could indicate that 30 entities simultaneously disappear every 900 frames, but this seems unlikely. The more plausible explanation is that all entity UID's are reset every 900 frames. After a short investigation, it was found that the labeling was done in segments of 900 frames and the UID's were not correctly stitched between segments. The MONCE metrics immediately revealed this issue and allowed the team to resolve it quickly.

\section{CONCLUSION}
MONCE is a performance evaluation methodology that expands on previous tracking benchmarks to assess long-term, non-contiguous, multi-object and detection model assisted trackers.  Beyond a simple "report card" that provides overall performance results but little depth, MONCE users are given diagnostic insight into tracking recall and precision performance as a function of tracking sequence length.  To isolate and identify specific failure modes, MONCE also provides lower-level performance metrics related to track longevity, localization, absence detection and re-acquisition.  Taken collectively, MONCE helps users perceive and comprehend the multi-faceted performance considerations in MOT.

% References
\bibliography{main} % bibliography data in main.bib

\begin{thebibliography}{1}

\bibitem{Kristan2015}
Kristan, M., Matas, J., Leonardis, A., Felsberg, M., Cehovin, L., Fernandez,
  G., Vojir, T., Hager, G., Nebehay, G., Pflugfelder, R., Gupta, A., Bibi, A.,
  Lukezic, A., Garcia-Martin, A., Saffari, A., Petrosino, A., and Montero,
  A.~S., ``The visual object tracking vot2015 challenge results,'' in [{\em
  2015 IEEE International Conference on Computer Vision Workshop
  (ICCVW)}{\nolinebreak\hspace{0.1em}]},   564--586 (2015).

\bibitem{Kristan2016}
Kristan, M., Matas, J., Leonardis, A., Vojir, T., Pflugfelder, R., Fernandez,
  G., Nebehay, G., Porikli, F., and Cehovin, L., ``A novel performance
  evaluation methodology for single-target trackers,'' {\em IEEE Transactions
  on Pattern Analysis and Machine Intelligence}~{\bf 38},  2137–2155 (Nov
  2016).

\bibitem{Wu2015}
Wu, Y., Lim, J., and Yang, M.-H., ``Object tracking benchmark,'' {\em IEEE
  Transactions on Pattern Analysis and Machine Intelligence}~{\bf 37}(9),
  1834--1848 (2015).

\bibitem{Benardin2008}
Benardin, K. and Stiefelhagen, R., ``Evaluating multiple object tracking
  performance: The clear mot metrics,'' {\em EURASIP}~{\bf 2208},  10 (2008).

\bibitem{Luiten2020}
Luiten, J., Osep, A., Dendorfer, P., Torr, P. H.~S., Geiger, A.,
  Leal{-}Taix{\'{e}}, L., and Leibe, B., ``{HOTA:} {A} higher order metric for
  evaluating multi-object tracking,'' {\em CoRR}~{\bf abs/2009.07736} (2020).

\bibitem{Luo2017}
Luo, W., Xing, J., Milan, A., Zhang, X., Liu, W., Zhao, X., and Kim, T.-K.,
  ``Multiple object tracking: A literature review,'' (2017).

\bibitem{Valmadre2018}
Valmadre, J., Bertinetto, L., Henriques, J.~F., Tao, R., Vedaldi, A.,
  Smeulders, A., Torr, P., and Gavves, E., ``Long-term tracking in the wild: A
  benchmark,'' (2018).

\bibitem{Lukezic2018}
Lukezic, A., Zajc, L.~C., Vojir, T., Matas, J., and Kristan, M., ``Now you see
  me: evaluating performance in long-term visual tracking,'' (2018).

\bibitem{Lukezic2016}
Lukezic, A., Vojir, T., Zajc, L.~C., Matas, J., and Kristan, M.,
  ``Discriminative correlation filter with channel and spatial reliability,''
  {\em CoRR}~{\bf abs/1611.08461} (2016).

\end{thebibliography}
\bibliographystyle{styles/spiebib} % makes bibtex use spiebib.bst

\end{document}